\documentclass[runningheads]{llncs}
\usepackage{graphicx}
\usepackage{amsmath}
\usepackage{cleveref}
\usepackage{multicol}
\usepackage{multirow}
\usepackage{xcolor}
\usepackage{booktabs}
\usepackage{subcaption}
\usepackage{algorithm}
\usepackage{algorithmic}

\usepackage[nocompress]{cite}
\usepackage{amsfonts}

\usepackage{pifont}
\begin{document}
\title{Metrics for saliency map evaluation of deep learning explanation methods\thanks{Supported by Nantes Excellence Trajectory (NExT)}}
%
%
\author{Tristan Gomez\inst{1} \and
Thomas Fréour\inst{2}\and Harold Mouchère\inst{1}
}
\authorrunning{T. Gomez et al.}
%
\institute{Nantes Université, Centrale Nantes, CNRS, LS2N, F-44000 Nantes, France 
\email{tristan.gomez@univ-nantes.fr};
\email{harold.mouchere@univ-nantes.fr}
\and
Nantes University Hospital, Inserm, CRTI, Inserm UMR 1064, F-44000 Nantes, France
\email{thomas.freour@chu-nantes.fr}\\
}
\maketitle              
\begin{abstract}
Due to the black-box nature of deep learning models, there is a recent development of solutions for visual explanations of CNNs. 
Given the high cost of user studies, metrics are necessary to compare and evaluate these different methods.
In this paper, we critically analyze the Deletion Area Under Curve (DAUC) and Insertion Area Under Curve (IAUC) metrics proposed by Petsiuk et al. (2018).
These metrics were designed to evaluate the faithfulness of saliency maps generated by generic methods such as Grad-CAM or RISE.
First, we show that the actual saliency score values given by the saliency map are ignored as only the ranking of the scores is taken into account.
This shows that these metrics are insufficient by themselves, as the visual appearance of a saliency map can change significantly without the ranking of the scores being modified.
Secondly, we argue that during the computation of DAUC and IAUC, the model is presented with images that are out of the training distribution which might lead to unexpected behavior of the model being explained.
To complement DAUC/IAUC, we propose new metrics that quantify the sparsity and the calibration of explanation methods, two previously unstudied properties.
Finally, we give general remarks about the metrics studied in this paper and discuss how to evaluate them in a user study.
\end{abstract}

\section{Introduction} 

Recent years have seen a surge of interest in interpretable machine learning, as many state-of-the-art learning models currently are deep models and suffer from their lack of interpretability due to their black-box nature.
In image classification, many generic approaches have been proposed to explain a model's decision by generating saliency maps that highlight the important areas of the image concerning the task at hand \cite{gradcam,gradcampp,rise,scoreCAM,varGrad,smoothGrad,liftcam,aUnifApproach}.
The community of interpretable deep learning has yet to find a consensus about how to evaluate these methods, the main difficulty residing in the ambiguity of the concept of interpretability. 
Indeed, depending on the application context, the users' requirements in terms of interpretability may vary a lot, making it difficult to find a universal evaluation protocol. 

This has started a trend in literature where authors confront users with models' decisions along with explanations to determine the users' preference on a particular application \cite{evalSalMap,evalVis,evalXAI}. 
The main issues of this approach are its financial cost and the difficulty to establish a correct protocol, which mainly comes from the requirement to design an experiment whose results will help understand the users' needs and also from the fact that most machine learning researchers are not used to run experiments involving humans.

Because of these issues, another trend proposes to design objective metrics to evaluate generic explanation methods \cite{rise,liftcam,gradcampp}.
In this paper, we chose to follow this trend, by proposing three new metrics.

We focus our work on the DAUC and IAUC metrics proposed by \cite{rise}.
First, we study several aspects of these metrics and we show that the actual saliency score values given by the saliency map are ignored as they only take into account the ranking of the scores. 
This shows that these metrics are insufficient by themselves, as the visual appearance of a saliency map can change significantly without the ranking of the scores being modified.
We also argue that during the computation of DAUC and IAUC, the model is presented with images that are out of the training distribution which might lead to unexpected behavior of the model and of the method used to generate the saliency maps.
We then introduce a new metric called Sparsity, which quantifies the sparsity of a saliency map, a property that is ignored by previous work.
Another property that was not studied until now is the calibration of the saliency maps.
Given it could be a useful property for interpretability, we also propose two new metrics to quantify it, namely Deletion Correlation (DC) and Insertion Correlation (IC).
Finally, we give general remarks about all the metrics studied in this paper and discuss how to evaluate these metrics in a user study.

\section{Existing metrics}

Various metrics have been proposed to automatically evaluate saliency maps generated by explanation methods \cite{liftcam,rise,gradcampp}.
These metrics consist to add or remove the important areas according to the saliency map and measure the impact on the initially predicted class score. 
For example, Chattopadhay et al. proposed ``increase in confidence'' (IIC) and ``average drop'' (AD) \cite{gradcampp}.
These metrics consist to multiply the input image with an explanation map to mask the non-relevant areas and to measure the class score variation.
Jung et al. proposed a variant of AD where the salient areas are masked instead of the non-salient, called Average Drop in Deletion (ADD) \cite{liftcam}.
In parallel, Petsiuk et al. proposed DAUC and IAUC which study the score variation while progressively masking/revealing the image instead of applying the saliency map once \cite{rise}. 
Given the similarity of these metrics, we will focus our study on DAUC and IAUC, which we will now describe.

\subsection{DAUC and IAUC }
To evaluate the reliability of the proposed attention mechanism, Petsiuk et al. proposed the Deletion Area Under Curve (DAUC) and Integration Area Under Curve (IAUC) metrics \cite{rise}. 
These metrics evaluate the reliability of the saliency maps by progressively masking/revealing the image starting with the most important areas according to the saliency map and finishing with the least important.

The input image is a 3D tensor $I \in \mathbb{R}^{H\times W \times 3}$ and the saliency map is a 2D matrix $S\in \mathbb{R}^{H'\times W'}$ with a lower resolution, $H'<H$ and $W'<W$.
First, $S$ is sorted and parsed from the highest element to its lowest element.
At each element $S_{i'j'}$, we mask the corresponding area of I by multiplying it by a mask $M^k \in \mathbb{R}^{H\times W}$, where 

\begin{equation}
    M_{ij}^k= 
\begin{cases}
    0,& \text{if } i'r<i<(i'+1)r~\text{and}~j'r<j<(j'+1)r\\
    1,              & \text{otherwise,}
\end{cases}
\end{equation}

where $r= H/H' = W/W'$.
After each masking operation, the model $m$ runs an inference with the updated version of I, and the score of the initially predicted class  is updated, producing a new score $c_k$ : 
\begin{equation}
     c_k = m(I \cdot \prod_{\tilde{k}=1}^{\tilde{k}=k} M^{\tilde{k}} ),
\end{equation}

where $k\in \{1,...,H' \times W'\}$. Examples of input images obtained during this operation can be seen in \cref{fig:outOfDistr}.
Secondly, once the whole image has been masked, the scores $c_k$ are normalized by dividing them by the maximum $\underset{k}{max}~c_k$ and then plotted as a function of the proportion $p_k$ of the image that is masked.
The DAUC is finally obtained by computing the area under the curve (AUC) of this function.
The intuition behind this is that if a saliency map highlights the areas that are relevant to the decision, masking them will result in a large decrease of the initially predicted class score, which in turn will minimize the AUC.
Therefore, minimizing this metric corresponds to an improvement.

Instead of progressively masking the image, the IAUC metric starts from a blurred image and then progressively unblurs it by starting from the most important areas according to the saliency map.
Similarly, if the areas highlighted by the map are relevant for predicting the correct category, the score of the corresponding class (obtained using the partially unblurred image) is supposed to increase rapidly.
Conversely, maximizing this metric corresponds to an improvement.

\subsection{Limitations}

\begin{figure}
    \centering
    
    \begin{subfigure}[t]{0.49\textwidth}
        \includegraphics[width=\textwidth]{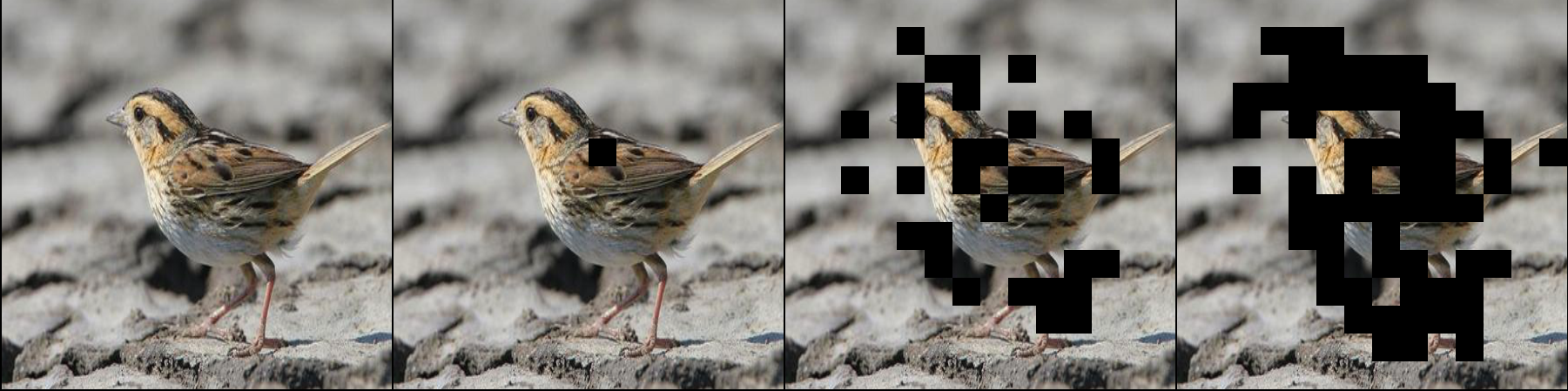}\\
        \includegraphics[width=\textwidth]{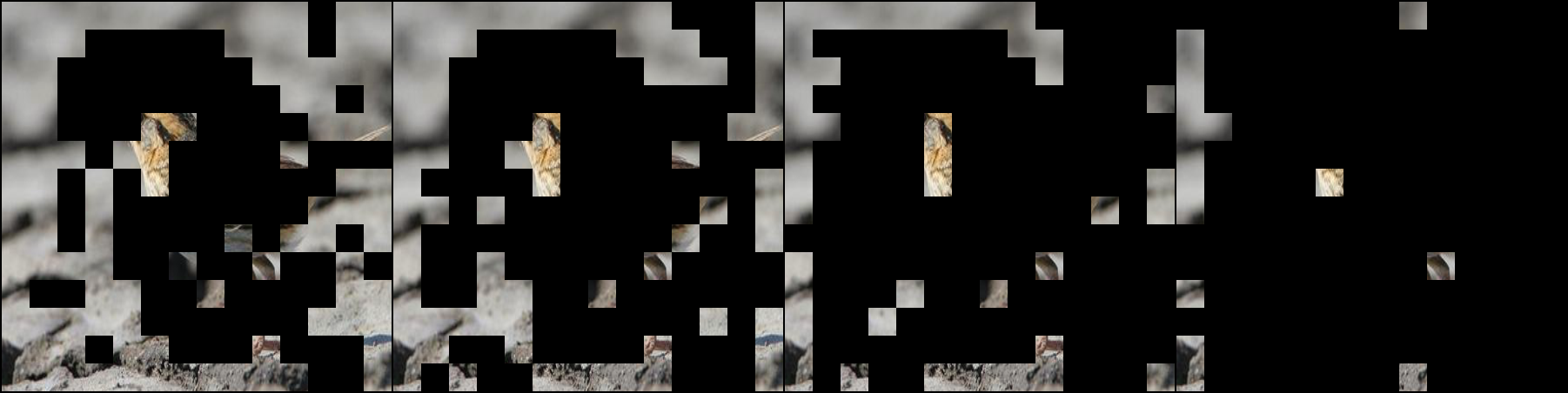}
        \caption{\label{OOD-DAUC}}
    \end{subfigure}
    \begin{subfigure}[t]{0.49\textwidth}
        \includegraphics[width=\textwidth]{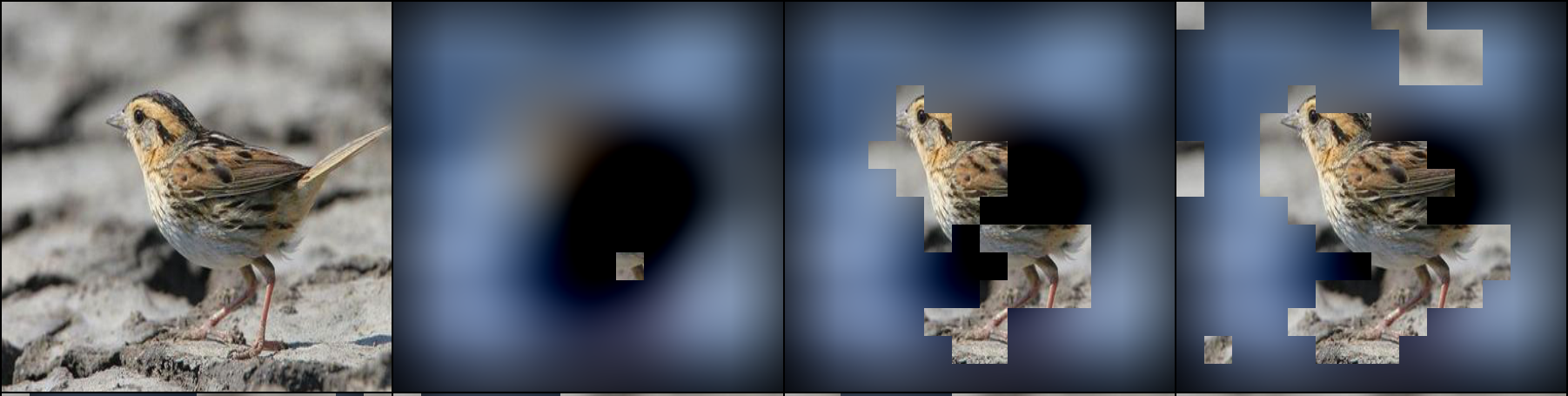}\\
        \includegraphics[width=\textwidth]{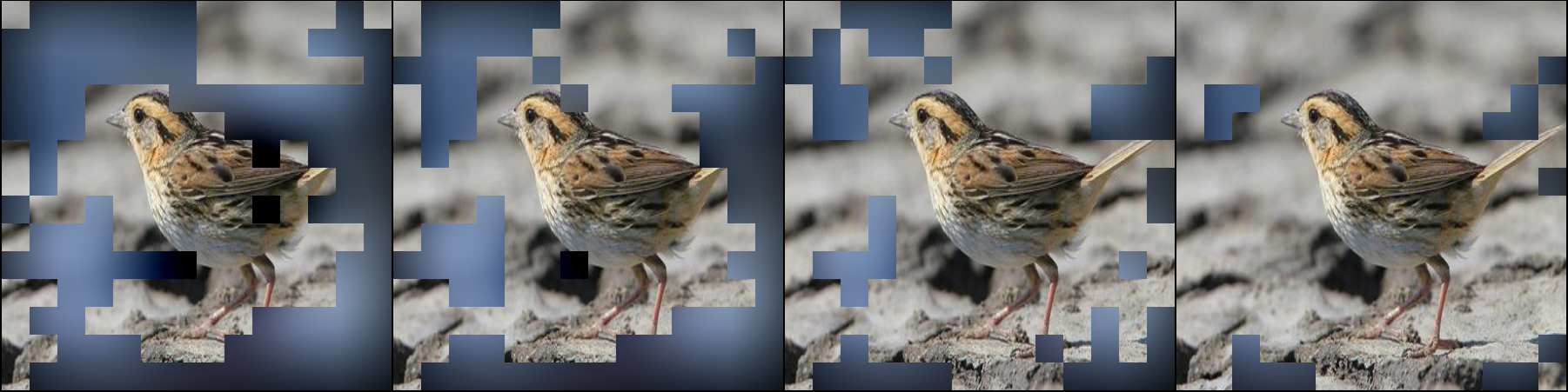}
        \caption{\label{OOD-IAUC}}
    \end{subfigure}    
    \caption{Examples of images passed to the model during the computation of (a) DAUC and (b) IAUC. Masking and blurring the input images probably lead to OOD  samples.}
    \label{fig:outOfDistr}
\end{figure}

\paragraph{\textbf{DAUC and IAUC generate out of distribution (OOD) images.}} When progressively masking/unblurring the input image, the model is presented with samples that can be considered out of the training distribution, as shown in \cref{fig:outOfDistr}.

Indeed, the kind of distortions produced by the masking/blurring operations do not exist naturally in the dataset and are  different from the kind produced by the standard data augmentations like random crop, horizontal flip, and color jitter, meaning that the model has not learned to process images with such distortions.
Therefore, the distribution of the images presented to the model is different from the one met during training.
However, it has been documented that CNNs and more generally deep learning models have poor generalization outside of the training distribution \cite{cnnOOD}.
This shows that DAUC and IAUC may not reflect the faithfulness of explanation methods as they are based on a behavior of the model that is different from that encountered when facing training distribution (e.g. during the test phase).

\begin{figure}
    \centering
    \begin{subfigure}[t]{0.45\textwidth}
        \includegraphics[width=\textwidth]{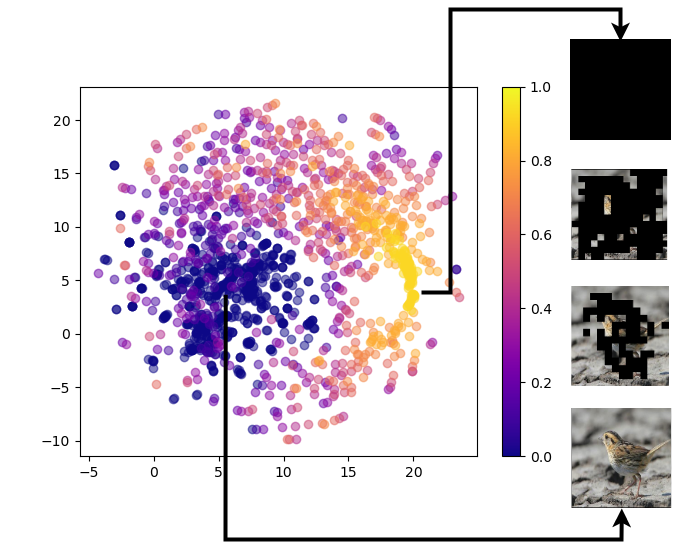}
        \caption{\label{DAUC_umap}}
    \end{subfigure}
    \begin{subfigure}[t]{0.45\textwidth}
        \includegraphics[width=\textwidth]{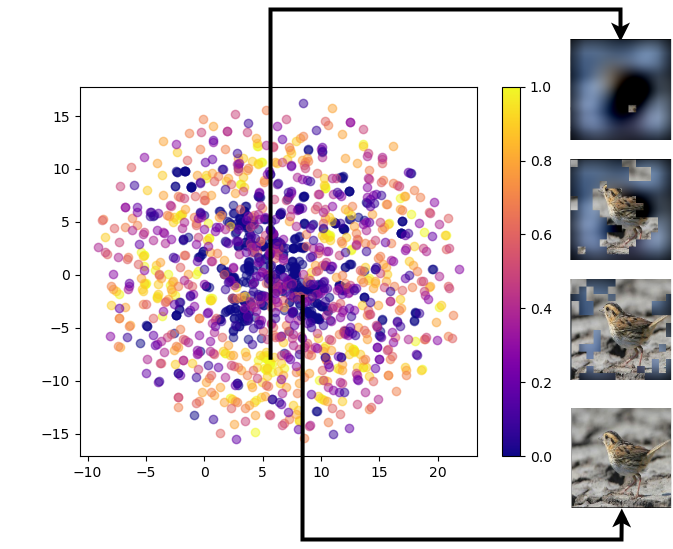}
        \caption{\label{IAUC_umap}}
    \end{subfigure}    
    \caption{UMAP projection of representations obtained while computing (a) DAUC and (b) IAUC on 100 images. 
    The color indicates the proportion of the image that is masked/unblurred. 
    The model used is a ResNet50 on which we applied Grad-CAM++ on the CUB-200-2011 dataset.
    We also plotted representations from 500 points of the test set to visualize the training distribution (in blue).
    By gradually masking the image, the representations converge towards a point (in yellow) that is distant from the points corresponding to unmasked images (in blue).
    Similarly, blurring the image causes the representation to move away from the training distribution.
    This shows that masking/blurring indeed creates OOD samples.
    }
    \label{fig:umap}
\end{figure}

To verify this hypothesis we visualize the UMAP \cite{umap} projections of the representations of 100 masked/blurred samples obtained during the computation of DAUC and IAUC on the CUB-200-2011 dataset \cite{CUB}. 
We also added the representation of 500 unmodified test images (in blue) to visualize the training distribution. 
The model used is a ResNet50 \cite{resnet} on which we applied Grad-CAM++ \cite{gradcampp}.
\Cref{fig:umap} shows that, during computation of DAUC, the representations gradually converge towards a unique point, which is not surprising as, at the end of the computation, all images are fully masked, i.e. plain black. 
However even when only a proportion of 0.4 of the image is masked, the corresponding representation is distant from the blue point cloud indicating the training distribution.
A similar phenomenon happens with IAUC, where blurring the image causes the representation to move away from the training distribution.
This experiment demonstrates that the DAUC and IAUC metrics indeed present OOD samples, which might lead to unexpected behavior of the model and of the method used to generate the explanation maps. 
However, as suspected by \cite{rise}, the blurring operation seems to create samples that are less far from the training distribution compared to the masking operation, probably because a blurred image still contains the low-frequency parts of the original image.
Another explanation is that most current classification models are designed with the assumption that an input image contains an object to recognize, which is in contradiction with DAUC and IAUC as they consist to remove the object to recognize from the image.
This suggests that modifying these metrics in such a way as to always leave an object to recognize in the input image would solve this issue. 

\paragraph{\textbf{DAUC and IAUC only take the pixel score rank into account.}} When computing DAUC and IAUC,
the saliency map is used only to determine in which order to mask/reveal the input image.
Hence, only the ranking of the saliency scores $S_{ij}$ is used to determine in which order to mask the image, leaving the actual values of the scores ignored. 

\begin{figure}[!hptb]
    \centering
    \begin{subfigure}[!hptb]{0.7\textwidth}
        \includegraphics[width=\textwidth]{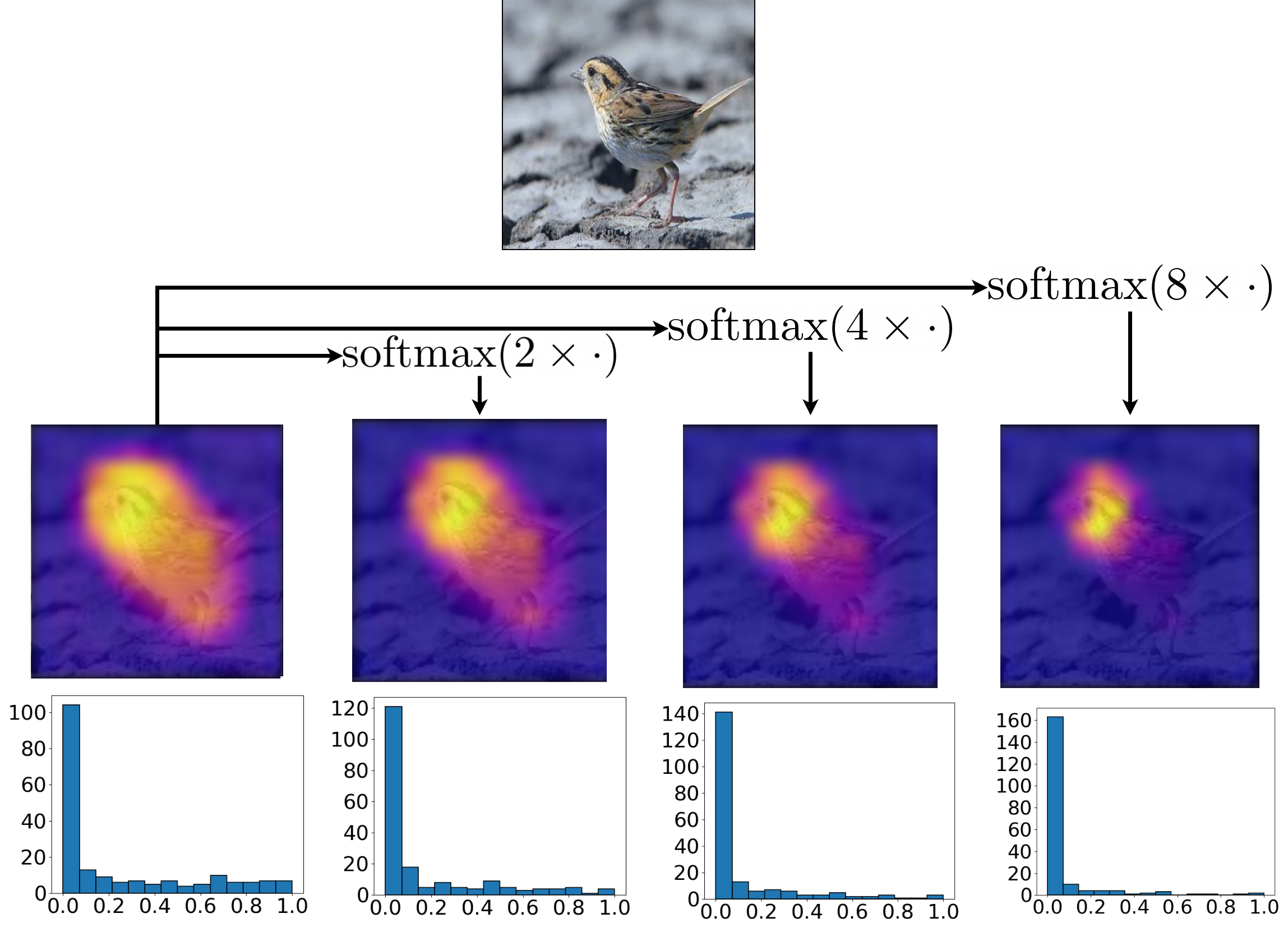}
        \caption{\label{introsalmap}}
    \end{subfigure}
     \begin{subfigure}[!hptb]{\textwidth}
        \centering
        \begin{tabular}{cccccccc}
            \toprule 
            Model & Viz. Method & Transformation        & DAUC   & IAUC   & Sparsity  & DC & IC \\
            \midrule
            \multirow{4}{*}{ResNet50}  & \multirow{4}{*}{Score-CAM}          & None                  &\multirow{4}{*}{0.012}  & \multirow{4}{*}{0.52}          & 3.98 & 0.187 & 0.22 \\
               &           &$\textrm{softmax}(2 \times \cdot)$ &  &     &           5.96 & 0.24 & 0.19 \\
               &           &$\textrm{softmax}(4 \times \cdot)$ &  &     &           8.96 & 0.29 & 0.14 \\
               &           &$\textrm{softmax}(8\times \cdot)$ &  &     &           16.52& 0.38 & 0.04 \\
               \bottomrule
        \end{tabular}
        \caption{\label{sparsity_intro_table}}
    \end{subfigure}
    \caption{Examples of saliency maps obtained by artificially sparsifying the saliency scores.
    The original saliency map is generated using Score-CAM applied on a ResNet50 model tested on the CUB-200-2011 dataset.
    Despite having different visual appearances, the four maps have the same DAUC and IAUC metric values because these metrics ignore the score values and only take into account the ranking of the scores. 
    On the other hand, the Sparsity metric depends on the score distribution and reflects the amount of focus of the map.
    In this figure, only the saliency map is modified, the decision process is left unchanged.
    \label{sparsity_intro}
    }
\end{figure}

However, pixel ranking is not the only characteristic that should be taken into account, as the visual appearance can vastly vary between two attention maps without changing the ranking. \Cref{sparsity_intro} shows examples of a saliency map with various score distributions artificially modified. We used a saliency map produced by the Score-CAM \cite{scoreCAM} explanation method and altered its score distribution by multiplying all values by a coefficient followed by the application of a softmax function. By increasing the coefficient we alter the visual appearance of the maps, without changing the pixel ranking, which maintains the same DAUC and IAUC scores. 
This illustrates the fact that DAUC and IAUC ignore the score dynamic of the saliency map, which can vastly affect the visual appearance.
To complement DAUC and IAUC, we propose new metrics that take the score values into account in the following section.

\section{Score aware metrics}

As mentioned in the previous paragraph, DAUC and IAUC ignore the actual score values and only take into account the saliency score $S_{ij}$ ranking.
To complement these metrics, we propose three new metrics, namely Sparsity, Deletion Correlation (DC), and Insertion Correlation (IC).

\subsection{The Sparsity metric}

An important visual aspect of saliency maps that has not been studied until now by the community is what we call sparsity. 
As shown by \cref{sparsity_intro}, Saliency maps can be more or less focused on a specific point depending on the score distribution, without changing the score ranking. 
This aspect could impact the interpretability of the method, as it significantly changes the visual aspect of the map and therefore could also affect the perception of the user.
For example, one could argue that a high sparsity value implies a map with a precise focus that highlights only a few elements of the input image, making it easier to understand for humans.
The Sparsity metric is defined as follows: 

\begin{equation}
\label{sparsity}
\textrm{Sparsity} = \frac{S_{max}}{S_{mean}}
\end{equation}

Where $S_{max}$ and $S_{mean}$ are respectively the maximum and mean score of the saliency map S. Note that the saliency methods available in the literature generate saliency maps with scores that are comprised in a large number of ranges. Therefore, the map should be first normalized as follows: 

\begin{equation}
    S' = \frac{S-S_{min}}{S_{max}-S_{min}}
\end{equation}

This means that, after normalization, $S'_{max}=1$ and \cref{sparsity} can be simplified to 
\begin{equation}
\textrm{Sparsity} = \frac{1}{S'_{mean}}    
\end{equation}

A high sparsity value means a high $S_{max}/S_{mean}$ ratio, i.e., a low mean score $S_{mean}$ which indicates that the map's activated areas are narrow and focused. 
As shown by \cref{sparsity_intro_table}, this metric is indeed sensitive to the actual saliency scores values and reflects the various amount of focus observed in the saliency maps.

\subsection{The DC and IC metrics}

As previously mentioned, the DAUC and IAUC metrics ignore the score values of the saliency maps and only take into account the ranking of the scores. 
This means that these metrics ignore the sparsity of the map, which is why we proposed to quantify this aspect.
Another potentially interesting property of saliency maps is the calibration.
The concept of calibration has seen a recent surge of interest in the deep learning community \cite{nncalibration,mixNMatch,measuringCalib}, but previous work focused exclusively  on calibrating prediction scores. 
A pixel $S_{ij}$ from a well-calibrated saliency map $S$ would reflect through its luminosity the importance it has on the class score.
More precisely, we say that an explanatory map $S$ is perfectly calibrated if for any two elements $S_{ij}$ and $S_{i'j'}$, we have $S_{ij}/S_{i'j'}=v/v'$, where $v$ and $v'$ are respectively the impact of $S_{ij}$ and $S_{i'j'}$ on the class score. 
To evaluate this, we propose to quantify how correlated the saliency scores and their corresponding impact on the class score are. To the best of our knowledge, this is the first time that an objective metric is proposed to measure the calibration of explanation methods.
In practice, such a metric could be used in a user study to evaluate to what extent the calibration property is useful.

We take inspiration from the DAUC and IAUC metrics and propose to gradually mask/reveal the input image by following the order suggested by the saliency map, but instead of computing the area under the class score vs. pixel rank curve, we compute the linear correlation of the class score variations and the saliency scores.
The correlation measured when masking the image is called Deletion Correlation (DC) and the one measured when revealing the image is called Insertion Correlation (IC).
The following paragraph details the computation of these two metrics.

DC is computed using the same progressive masking and inference method as DAUC. 
Once the scores $c_k$ have been computed, we compute the variation of the scores $v_k = c_k - c_{k+1}$.
Finally, we compute the linear correlation between the $v_k$ and the $s_k$ where $s_k$ is the saliency score of the area masked at step k.
For the IC metric, we take inspiration from IAUC, and instead of masking the image, we start from a  blurred image, and gradually reveal the image according to the saliency map. 
Once the image is totally revealed, the score variations are computed $v_k = c_{k+1}-c_k$ and we compute the linear correlation of the $v_k$ with the $s_k$.
Note that the order of the subtraction is reversed compared to DC because when revealing the image, the class score is expected to increase.

When computing DC/IC on a well-calibrated saliency method, we expect that when the class score variation is high, the saliency score should also be high, and conversely, when the class score variation is low, the saliency score should be also low.

The DC and IC metrics measure the calibration, which is an aspect that is ignored by the DAUC and IAUC metrics but also by the Sparsity metric. 
To illustrate, the DC and IC metrics are computed for the examples visible in \cref{sparsity_intro}.

\subsection{Limitations}

\paragraph{\textbf{The Sparsity metric does not take into account the prediction scores.}} Indeed, this metric only considers the saliency score dynamic and ignores the class score produced by the model. 
However, this is not necessarily a problem as this metric was designed to be used as a complement to other metrics like DAUC, IAUC, DC, or IC, which takes the class score into account.

\paragraph{\textbf{The DC and IC metrics also generate OOD images.}} As we took inspiration from DAUC and IAUC and also passed masked/blurred examples to the model, one can make the same argument as for DAUC/IAUC to show that the reliability of DC and IC could probably be improved by preventing OOD samples.

\section{Benchmark\label{secbench}}

We compute the five metrics studied in the work (DAUC, IAUC, DC, IC, and Sparsity) on post-hoc generic explanation methods and attention architectures that integrate the computation of the saliency map in their forward pass.
The post-hoc methods are Grad-CAM \cite{gradcam}, Grad-CAM++ \cite{gradcampp}, RISE \cite{rise}, Score-CAM \cite{scoreCAM}, Ablation CAM \cite{ablationCAM}. 
The  architectures with native attention are B-CNN \cite{seeBetterBeforeLooking}, BR-NPA \cite{brnpa}, the model from \cite{interByParts} which we call IBP (short for Interpretability By Parts), ProtoPNet \cite{ProtoPNet}, and ProtoTree \cite{prototree}.
These attention models generate several saliency maps (or \textit{attention} maps) per input image but the metrics are designed for a single saliency map per image. 
To compute the metrics on these models, we selected the first attention map among all the ones produced, as, in these architectures, the first is the most important one.

\Cref{benchmark} shows the performances obtained. The most important thing to notice is the overall low values of correlation, especially for IC, where most values are very close to 0, meaning the saliency scores reflect the impact on the class score as much as random values.
This highlights the fact that attention models and explanation methods are currently not designed for this objective, although it could be an interesting property.

\begin{table}[]
\centering
\noindent\makebox[\textwidth]{
\begin{tabular}{c|c|c ccccc}
\toprule 
Model&Viz. Method&Accuracy&DAUC&IAUC&DC&IC&Sparsity\\ 
\midrule 
\multirow{6}{*}{ResNet50}&Ablation CAM&\multirow{6}{*}{0.842}&$0.0215$&$0.26$&$0.36$&$-0.04$&$8.54$\\ 
&Grad-CAM&&$0.0286$&$0.16$&$0.35$&$-0.12$&$5.28$\\ 
&Grad-CAM++&&$0.0161$&$0.21$&$0.35$&$-0.07$&$6.73$\\ 
&RISE&&$0.0279$&$0.18$&$\mathbf{0.57 }$&$-0.11$&$6.63$\\ 
&Score-CAM&&$0.0207$&$0.27$&$0.32$&$-0.05$&$5.96$\\ 
&AM&&$0.0362$&$0.22$&$0.31$&$-0.09$&$4.04$\\ 
\hline 
B-CNN&\multirow{5}{*}{-}&0.848&$0.0208$&$0.3$&$0.27$&$-0.02$&$12.74$\\ 
BR-NPA&&$\mathbf{0.855 }$&$0.0155$&$\mathbf{0.49 }$&$0.41$&$-0.02$&$\mathbf{16.02 }$\\ 
IBP&&0.819&$0.0811$&$0.48$&$0.23$&$-0.04$&$6.56$\\ 
ProtoPNet&&0.848&$0.2964$&$0.37$&$0.1$&$-0.06$&$2.18$\\ 
ProtoTree&&0.821&$0.2122$&$0.43$&$0.17$&$\mathbf{0.04 }$&$13.75$\\ 
\bottomrule
\end{tabular}}
\caption{Evaluation of the interpretability on the CUB-200-2011 dataset.}
\label{benchmark}
\end{table}

\section{Discussion}

One common limit of all the metrics discussed here is that they are designed for methods and architectures producing single saliency maps, making their use for multi-part attention architectures like B-CNN, BR-NPA, IBP, ProtoPNet, and ProtoTree less straightforward. 
In \cref{benchmark}, we chose to only select the most important attention map but the other ones should also be taken into account to fully reflect the model's behavior. 
We also could have computed the mean attention maps from all the ones produced by the model but this would also not be faithful towards the model.
Indeed, that would amount to considering that all attention maps have the same weight in the decision, which is not true, as the first attention map has more importance in the decision than the second, which is more important than the third, etc.
One possibility would be to estimate the weight of each map and to compute a pondered mean but there remains the issue of computing the weight, which may be difficult in the general case, due to the variety of architectures.


Note that the low values of DC and IC in \cref{benchmark} do not imply that the models and methods provide unsatisfying performance, but simply show that the calibration property has not been studied until now.

We propose to quantify the sparsity and the calibration of the saliency maps as these properties have not been studied until now and may be relevant for interpretability.
However, to what extent this is true remains to be tested in a subjective experiment. More generally, all the metrics discussed in this paper should be tested against a user experiment. As shown by \cite{whatArePeople,humanXAI}, there is a great variety of possible experimental setups depending on what should be explained, in which context, and for who the explanation is targeted.

Notably, how the explanation is presented to the user is still an open question. 
For example, \cite{reliablePostHOC} applies a mask on the input image whereas \cite{evalSalMap} proposed to superimpose the explanation over the image.
Also, various tasks can be given to the user to evaluate the explanation.
Slack et al. proposed to mask the input image at the most salient areas and ask users to try to recognize the object with the mask \cite{reliablePostHOC}.
Instead of guessing the label, Alqaraawi et al. asked users to predict the network's prediction to evaluate if the access to a saliency map helps to improve their prediction \cite{evalSalMap}.
We could use a similar setup with various saliency map methods and evaluate the users' accuracy.
Then we could rank the methods according to the impact they have on the users' accuracy and see how this ranking relates to the ranking provided by the objective metrics.
Like this, we could deduce which metrics best reflect the  impact of an explanation method on the user's understanding of the model.

The main issue is that current user studies seem to show that providing saliency maps to the user affects little their understanding of the model and also does not affect the trust in the model \cite{expAI}.
One study found that the presence of a saliency map helps users to better predict the model's output \cite{evalSalMap}.
However, the effect size measured was small and it could be argued that the effect of changing the explanation method would be smaller or even difficult to observe.

\section{Conclusion}

In this work, we first studied two aspects of the DAUC and IAUC metrics.
We showed that they may generate OOD samples which might negatively impact their reliability.
Also, we show that they only take into account the ranking of the saliency scores and show that the visual appearance of a saliency map can significantly change without the DAUC and IAUC metrics being affected.
Then, we propose to quantify two aspects that were previously unstudied on saliency maps, the sparsity and the calibration (DC and IC).
Finally, we conclude with general remarks on the studied metrics and discuss the issues of a user study that could be used to evaluate the usefulness of these metrics.

\bibliographystyle{splncs04}
\bibliography{biblio}

\end{document}